\pdfoutput=1
\documentclass[11pt]{article}

\usepackage[]{ACL2023}
\usepackage{tablefootnote}
\usepackage{times}
\usepackage{latexsym}
\usepackage{graphicx}
\usepackage{url}
\usepackage{enumerate}
\usepackage[T1]{fontenc}
\usepackage{multirow}
\usepackage{multicol}
\usepackage[utf8]{inputenc}
\usepackage{microtype}
\usepackage{inconsolata}

\title{CHEAT: A Large-scale Dataset for Detecting ChatGPT-writtEn AbsTracts}

\author{Peipeng Yu \\
  Jinan University \\
  \texttt{ypp865@163.com} \\\And
  Jiahan Chen \\
  Jinan University \\
  \texttt{335486990@qq.com} \\\And
  Xuan Feng \\
  Jinan University \\
  \texttt{fenffef@163.com} \\\And
  Zhihua Xia \\
  Jinan University \\
  \texttt{xia\_zhihua@163.com} \\}
  
\begin{document}
\maketitle

\begin{abstract}

	The powerful ability of ChatGPT has caused widespread concern in the academic community. Malicious users could synthesize dummy academic content through ChatGPT, which is extremely harmful to academic rigor and originality. The need to develop ChatGPT-written content detection algorithms call for large-scale datasets. In this paper, we initially investigate the possible negative impact of ChatGPT on academia, and present a large-scale CHatGPT-writtEn AbsTract dataset (CHEAT) to support the development of detection algorithms. In particular, the ChatGPT-written abstract dataset contains 35,304 synthetic abstracts, with $Generation$, $Polish$, and $Mix$ as prominent representatives. Based on these data, we perform a thorough analysis of the existing text synthesis detection algorithms. We show that ChatGPT-written abstracts are detectable, while the   detection difficulty increases with human involvement. Our dataset is available in https://github.com/botianzhe/CHEAT.

\end{abstract}

\section{Introduction}

ChatGPT, a natural language processing tool based on artificial intelligence technology, has attracted widespread attention in recent times. Based on user needs, ChatGPT could complete tasks with high quality, such as coding, translation, thesis writing\cite{macdonald2023can}, and so on. While ChatGPT brings convenience to human life, the potential harm of its synthetic content has gradually emerged\cite{ufuk2023role}. As an important part of the academic thesis, the abstract has been proven to be brilliantly synthesized by ChatGPT\cite{gao2022comparing,abstract}. Malicious researchers could synthesize plausible academic content without practical research, which seriously undermines academic originality.\cite{science.adg7879}. Meanwhile, although the synthetic content exhibit high levels of authenticity, the academic rigor and the conclusion correctness cannot be guaranteed\cite{else2023abstracts}. How to detect ChatGPT-written abstracts and ensure the academic originality have been pressing issues that need to be addressed.

\begin{figure}[tbp]
	\centering
	\includegraphics[width=1.0\linewidth]{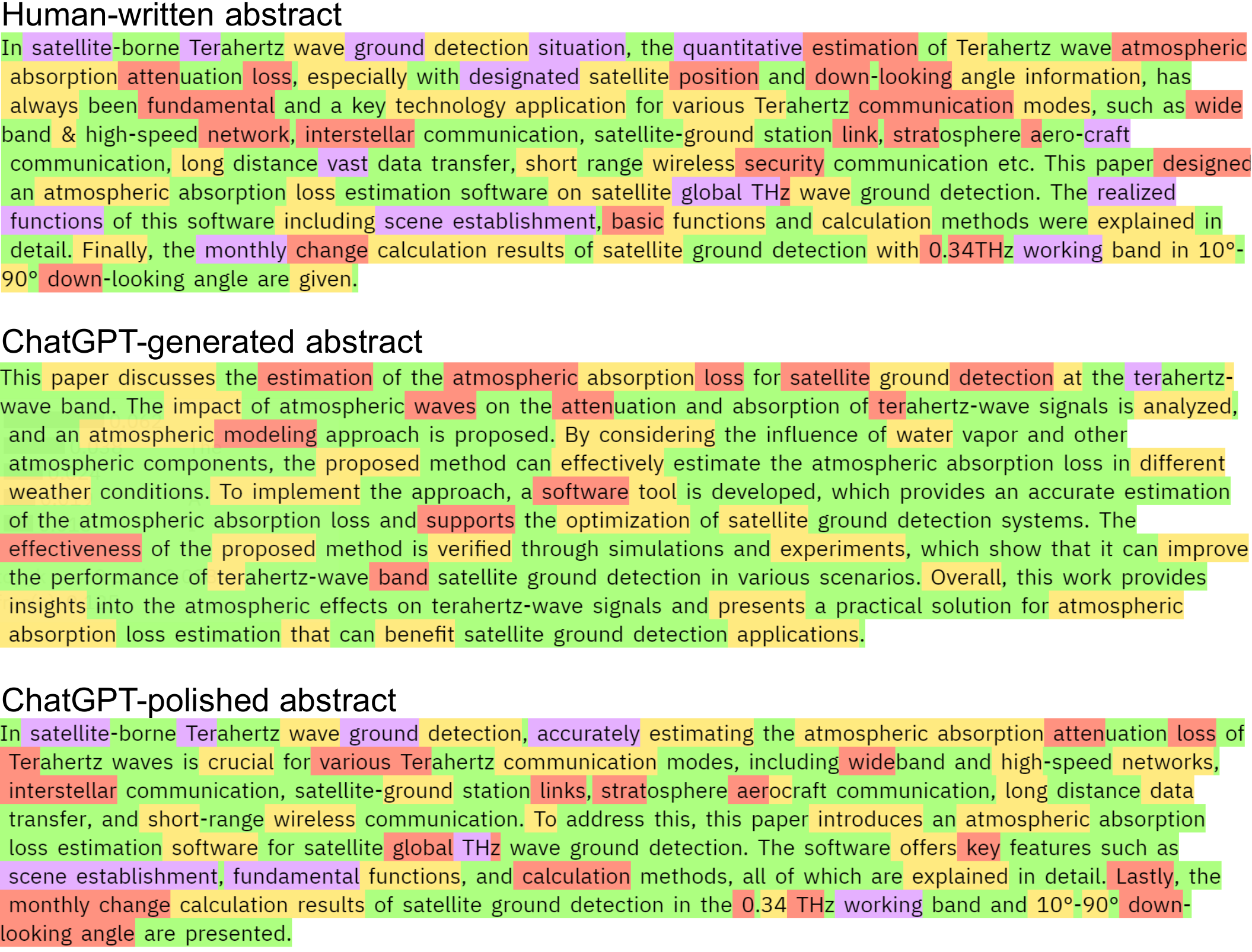}
	\centering
	\caption{The different distributions of human-written and ChatGPT-written abstracts. The visualization results are obtained by GLTR \cite{gehrmann2019gltr}.}
	\label{fig:example}
\end{figure}

As concerns over ChatGPT escalate, there has been a surge of interest in detecting ChatGPT content. The availability of large-scale datasets is an enabling factor for the development of ChatGPT content detection methods. So far, researchers have constructed multiple detection datasets for question answering and querying scenarios\cite{guo2023close,mitrovic2023chatgpt}. Some studies also evaluated the quality of ChatGPT-written text with a small amount of data\cite{huang2023chatgpt,cao2023assessing,gao2022comparing}. However, due to the limitations of OpenAI, there is currently no large-scale dataset to support the development of ChatGPT-written abstract detection algorithms.\footnote{https://www.nature.com/articles/d41586-023-00553-9}. 

In this paper, we focus on the practical need to generate large-scale and diverse ChatGPT-written abstracts. We explore different abstract synthesis methods using ChatGPT, including $Generation$, $Polish$, and $Mix$. As shown in Figure \ref{fig:example}, ChatGPT-written abstracts are almost indistinguishable from the human-written ones, but there are some differences in semantic distribution. We aim to construct a large-scale dataset to analyze the abnormal distribution patterns in ChatGPT-written abstracts. The contributions are summarized as follows:

\begin{enumerate}[(1)]
	\item We present a large-scale ChatGPT-written abstract dataset, CHEAT, which is currently the largest dataset available for ChatGPT-written abstract detection. 
	\item We analyze the distribution differences between human-written and ChatGPT-written abstracts. Compared with human-written abstracts, ChatGPT-written abstracts are more informative, but lack of logic and definition precision.

	\item We conduct an evaluation of current detection algorithms on the CHEAT dataset. The experimental results show that existing schemes lack effectiveness in detecting ChatGPT-written abstracts, and the detection difficulty increases with human involvement.

\end{enumerate}

\section{Related Work}

In this section, we will introduce the related works of Text synthesis and Text synthesis detection.
\subsection{Text Synthesis}

ChatGPT, an artificial intelligence chatbot program, is  developed based on the Generative Pre-trained Transformer (GPT) models. The earliest GPT-1\cite{radford2018improving} was proposed in 2018, which could perform various natural language processing tasks such as semantic inference, question answering, and classification. Although GPT-1 could effectively handle some unknown tasks, its generalization ability was still not sufficient compared to fine-tuned supervised models. After that, Radford \emph{et al.}\cite{radford2019language} proposed GPT-2 in 2019, which used more network parameters and larger datasets to learn more universal knowledge and achieved the best performance in multiple specific language modeling tasks. Later, Brown \emph{et al.} \cite{brown2020language} proposed GPT-3. They employed self-supervised mechanisms to learn general knowledge and performed well in various tasks. However, due to directly learning from large-scale text corpora, GPT-3 was prone to generating erroneous or offensive content, which limited its application. To solve this problem, Ouyang \emph{et al.}\cite{ouyang2022training} designed the InstructGPT language model to improve the quality of synthetic content. They combined supervised learning and Reinforcement Learning from Human Feedback(RLHF) to promote the synthetic content better following user intent. The recent ChatGPT furtherly learned to rank the quality of output results, and improved the model's understanding ability successfully. The Instruct Learning and Reinforcement Learning strategies had driven the success of ChatGPT, enabling it to synthesize academic contents that can deceive most detectors.

\subsection{Text Synthesis Detection}

Existing text synthesis detection algorithms could be roughly divided into traditional detection algorithms and deep learning-based detection algorithms. Traditional detection algorithms analyze  abnormal patterns of synthetic text by extracting hand-crafted features, while deep learning-based algorithms train detection models on large-scale datasets to achieve accurate detection of synthetic text. The following provides a detailed introduction to the two types of approaches.

\textbf{Traditional text synthesis detection} Early text synthesis algorithms left obvious synthesis traces and abnormal distribution patterns in statistical features. Sebastian \emph{et al.} \cite{gehrmann2019gltr} developed a visualization tool called GLTR. They integrated multiple detection methods based on statistical features(word order, predicted distribution entropy, and so on.) to identify anomalies in synthetic text. Fröhling \emph{et al.} \cite{frohling2021feature} found that most synthetic texts were repetitive and lacked purpose, so they extracted statistical features of text style and successfully identified various synthetic text using Random Forests and Support Vector Machines. Lundberg \emph{et al.} \cite{levin2023identifying} generated 50 abstracts using ChatGPT and then analyzed the differences between human-written abstracts and ChatGPT-written abstracts using Grammarly. They found that ChatGPT was able to generate more unique words and had fewer grammatical errors. Recently, Guo \emph{et al.} \cite{guo2023close} discovered that ChatGPT-written text tends to use more connecting words, providing new ideas for detecting synthetic text. 

\begin{table*}[!h]
	\caption{The searching keywords used for collecting human-written abstracts.}
	\resizebox{\linewidth}{!}{
		\begin{tabular}{|c|c|c|c|c|}
			\hline
			Natural language processing & Feature extraction & Artificial Intelligence & Knowledge Representation and   Reasoning & Internet of Things       \\ \hline
			Computational   modeling    & Labeling           & Machine Learning        & Expert Systems                           & Cloud Computing          \\ \hline
			Training                    & Neural networks    & Deep Learning           & Fuzzy Logic                              & Cybersecurity            \\ \hline
			Supervised   learning       & Nonlinear systems  & Computer Vision         & Genetic Algorithms                       & Data Mining              \\ \hline
			Brightness                  & Convergence        & Robotics                & Swarm Intelligence                       & Predictive Analytics     \\ \hline
			Estimation                  & networks           & Reinforcement Learning  & Big Data Analytics                       & Decision Support Systems \\ \hline
		\end{tabular}}
	\label{tab:keywords}
\end{table*}

\textbf{Deep learning-based text synthesis detection} The rich information stored in large language models could provide guidance for text synthesis detection. Solaiman \emph{et al.} \cite{solaiman2019release} fine-tuned the detection model based on RoBERTa and achieved the best performance in the generated-web detection task. Tay \emph{et al.} \cite{tay2020reverse} found that different language models would leave different defects in the synthesized text. Thus, they trained  corresponding detectors for specific language models to achieve high-accuracy detection, but performed poorly in cross-model detection. After that, Ippolito \emph{et al.} \cite{ippolito2020automatic} constructed a novel text synthesis dataset and fine-tuned the BERT classification model, significantly improving the detection accuracy on synthetic texts. Although deep learning algorithms could detect synthetic text well, their interpretability greatly limits their application. The research focus of current deep learning based solutions is to explain the underlying principles behind the detection results.

\section{The ChatGPT-written Abstract Dataset}

Although some datasets for ChatGPT-written content detection are available, there is no large-scale ChatGPT-written abstract datasets until now. To provide more relevant data for developing detection methods, we constructed the ChatGPT-written abstract (CHEAT) dataset. 

% Table * summarizes the comparison of our dataset with other existing datasets.

\subsection{Basic information}
The CHEAT dataset is consisted of 15,395 human-written abstracts and 35,304 ChatGPT-written abstracts. The average length of all abstracts is 163.9 and the total vocabulary size is 130,272. The human-written abstracts are searched from IEEE Xplore, a mega repository of scholarly literature. As shown in Table \ref{tab:keywords}, we select 30 keywords to search for matching article abstracts. The collected abstracts all originate from the field of computer science, and cover areas such as  natural language processing, computer vision, and machine learning. Among them, abstracts with less than 100 words accounted for 11.6\%, abstracts with 100-200 words accounted for 67.7\%, and abstracts with more than 200 words accounted for 20.7\%. ChatGPT-written abstracts are then synthesized through the interface provided by OpenAI.

\subsection{Abstract Synthesis Methods}
The synthetic abstracts in the CHEAT dataset are consisted of $Generation$ abstracts, $Polish$ abstracts, and $Mix$ abstracts. We input the human-written abstracts into ChatGPT (gpt-3.5-turbo) through the OpenAI interface to obtain the corresponding output. Specifically, we obtain synthesized abstracts in three ways:

\textbf{Generation:} ChatGPT is capable of generating plausible abstracts from keywords. Malicious users could use ChatGPT to generate abstracts directly for publication, thus obtaining scientific results without any cost. To detect such synthesis, we create the ChatGPT-Generation dataset to develop detection algorithms. Specifically, we use ChatGPT to output the generated abstract by entering the following command: "\emph{Generate a 200-word abstract of the paper in English based on the title and keywords; your answer only needs to include the generated paragraph.}", followed by the title and keywords in the human-written data. 

\textbf{Polish:} Unlike traditional text polishing algorithms, ChatGPT is able to utilize its rich linguistic knowledge to optimize the original text and enhance its readability. Malicious users are likely to evade paper checking by text polish, causing damage to academic originality\cite{khalil2023will}. In this paper, we create the ChatGPT-Polish dataset to support the relevant detection algorithm. Specifically, we employ ChatGPT to output a polished abstract by entering the following command: "\emph{Polish the following paragraphs in English, your answer just needs to include the polished text.}", followed by the human-written abstract.

\textbf{Mix:} Malicious users are likely to mix  human-written abstracts with polished abstracts to evade detection algorithms. To address this problem, we create a more challenging dataset, ChatGPT-Mix, based on the polished abstracts. Specifically, we first decompose the polished abstracts and human-written abstracts according to their semantics, then construct a random mask to determine which sentences need to be replaced, and finally replace the polished abstracts with text from the human-written abstracts to obtain the final mixed abstract. By controlling the number of 1 in the mask, we are able to effectively control the text replacement rate and synthesize mixed abstracts with different detection difficulties.

\section{Evaluation and Analysis}
ChatGPT's ability to synthesize abstracts is potentially harmful to academic originality and correctness. In this section, we first analyze the linguistic differences between human-written and ChatGPT-written abstracts, and then evaluate the detection performance of existing algorithms on the CHEAT dataset. After that, we explore the judgment basis of the deep learning based detection methods.

\subsection{Linguistic Analysis}
In this subsection, we evaluate the synthesis quality of ChatGPT-written abstracts from a linguistic perspective, including lexical analysis and dependency analysis.

\begin{figure}[tbp]
	\centering
	\includegraphics[width=1.0\linewidth]{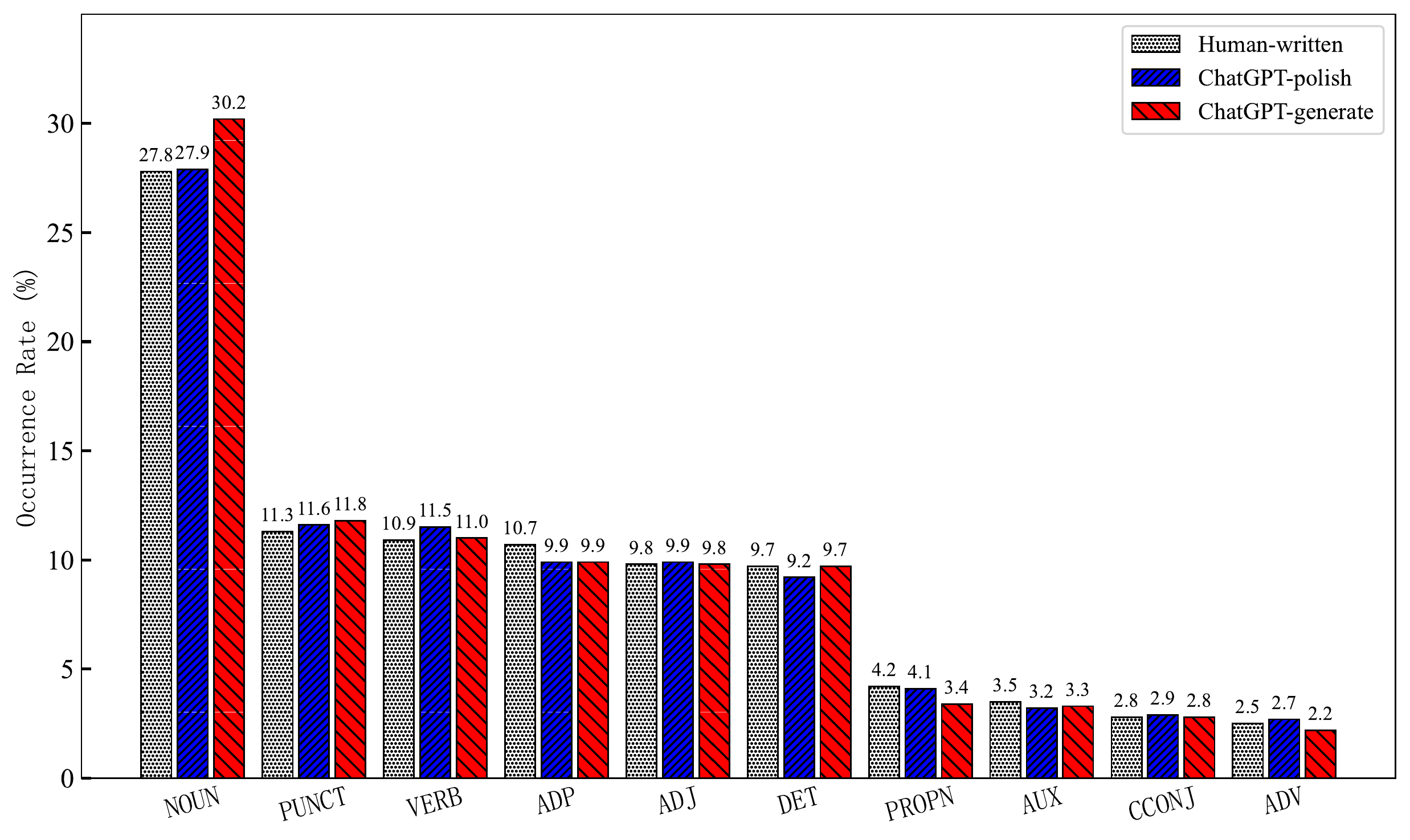}
	\centering
	\caption{The lexical distribution for human-written, ChatGPT-polished, and ChatGPT-generated abstracts.}
	\label{fig:vocabulary}
\end{figure}

\textbf{Lexical analysis}
In the field of natural language processing, each word  could be classified as one of lexical categories. The part-of-speech (POS) tagging task aims to determine the grammatical class of each word in a given sentence. In this part, we use the POS module in NLTK \cite{bird2006nltk} to calculate the lexical distributions of abstract texts in the CHEAT dataset, and sort them by lexical percentage. As shown in Figure \ref{fig:vocabulary}, we show the statistics of the top ten lexicalities. It can be seen that noun ($NOUN$) occupies the largest proportion of all lexicalities, while punctuation ($PUNCT$), verb ($VERB$), adposition ($ADP$), adjective ($ADJ$), and determiner ($DET$) occupy most of the others. Comparing human-written and ChatGPT-written abstracts, the following findings could be made:

(1) The proportions of $NOUN$, $VERB$ in ChatGPT-written abstracts are higher than those of human-written abstracts. It can be argued that the rich knowledge contained in ChatGPT can provide a more diverse vocabulary for abstract synthesis and make it more informative.

(2) The proportions of adposition($ADP$), proper noun($PROPN$), and auxiliary($AUX$) in human-written abstracts are larger than those of ChatGPT-written abstracts. This indicates that humans tend to pay more attention to abstract structure, consistency, and logic, while ChatGPT is weaker in these aspects.

\begin{figure}[tbp]
	\centering
	\includegraphics[width=1.0\linewidth]{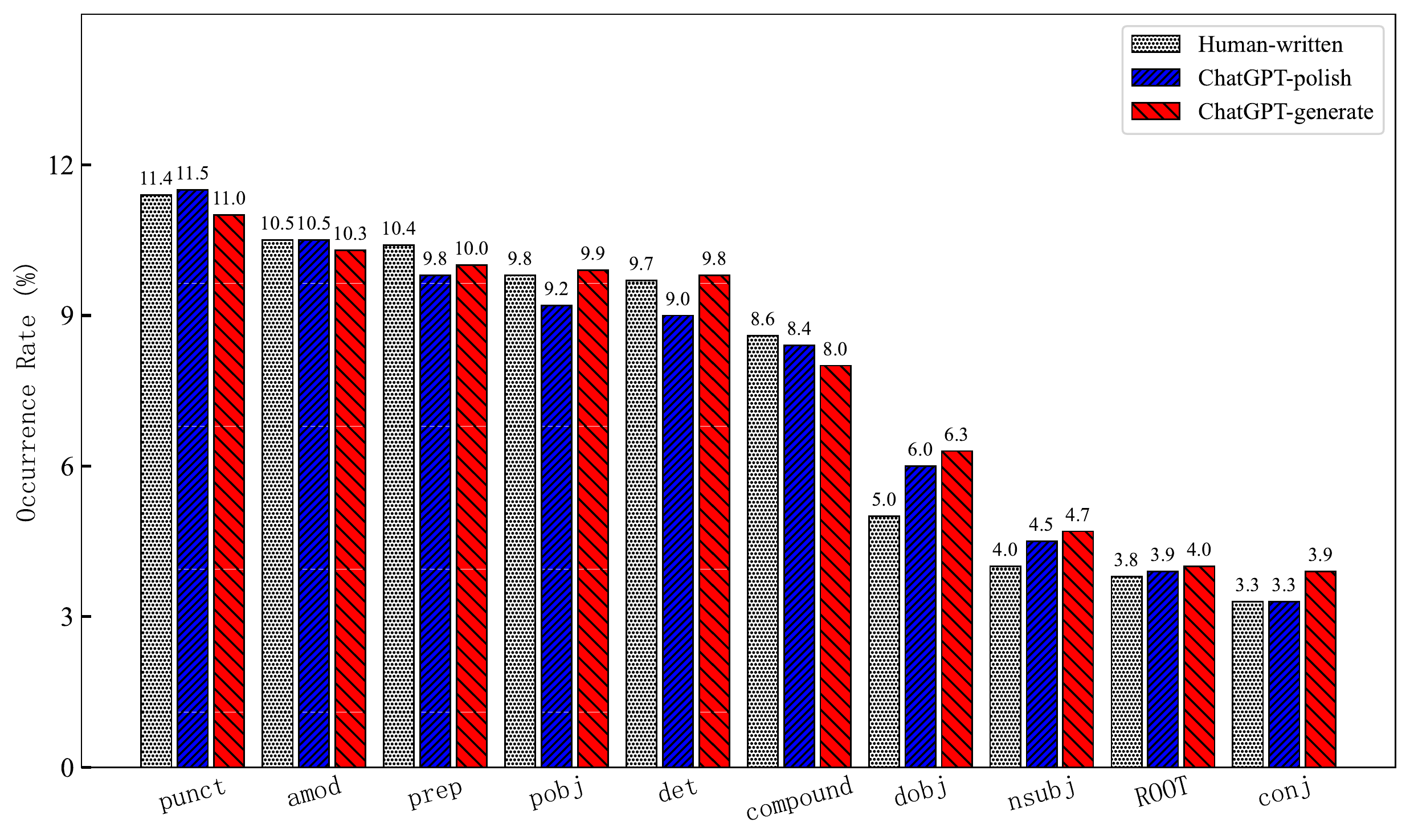}

	\caption{The dependency distribution forhuman-written, ChatGPT-polished, and ChatGPT-generated abstracts.}
	\label{fig:dependency}
\end{figure}
\begin{table*}[tbp]
	\centering
	\caption{The detection performance of existing schemes. The accuracy and AUC(Area Under the ROC Curve) are used as metrics.}
	\setlength{\tabcolsep}{2.6mm}{
		\begin{tabular}{|c|cccccc|}
			\hline
			\multirow{3}{*}{Methods}    & \multicolumn{6}{c|}{Datasets}                                                                                                                                                        \\ \cline{2-7}
			                            & \multicolumn{2}{c|}{Generation} & \multicolumn{2}{c|}{Polish}        & \multicolumn{2}{c|}{Mix}                                                                                     \\ \cline{2-7}
			                            & \multicolumn{1}{c|}{ACC}       & \multicolumn{1}{c|}{AUC}           & \multicolumn{1}{c|}{ACC}   & \multicolumn{1}{c|}{AUC}            & \multicolumn{1}{c|}{ACC}   & AUC            \\ \hline
			Grover \cite{zellers2019defending}                     & \multicolumn{1}{c|}{54.24}     & \multicolumn{1}{c|}{56.34}         & \multicolumn{1}{c|}{53.33} & \multicolumn{1}{c|}{55.45}          & \multicolumn{1}{c|}{50.89} & 51.71          \\ \hline
			Zerogpt\cite{AITextDetector}                    & \multicolumn{1}{c|}{67.32}     & \multicolumn{1}{c|}{78.8}          & \multicolumn{1}{c|}{52.71} & \multicolumn{1}{c|}{57.35}          & \multicolumn{1}{c|}{50.61} & 52.59          \\ \hline
			OpenAI-detector \cite{solaiman2019release}           & \multicolumn{1}{c|}{75.97}     & \multicolumn{1}{c|}{84.41}         & \multicolumn{1}{c|}{54.07} & \multicolumn{1}{c|}{56.17}          & \multicolumn{1}{c|}{52.18} & 55.23          \\ \hline
			ChatGPT-detector-roberta \cite{guo2023close}   & \multicolumn{1}{c|}{75.54}     & \multicolumn{1}{c|}{81.91}         & \multicolumn{1}{c|}{53.65} & \multicolumn{1}{c|}{47.28}          & \multicolumn{1}{c|}{51.92} & 63.71          \\ \hline
			Chatgpt-qa-detector-roberta \cite{guo2023close} & \multicolumn{1}{c|}{85.56}     & \multicolumn{1}{c|}{\textbf{97.6}} & \multicolumn{1}{c|}{53.53} & \multicolumn{1}{c|}{\textbf{64.39}} & \multicolumn{1}{c|}{51.67} & \textbf{65.28} \\ \hline
		\end{tabular}}
	\label{tab:compare_withouttrain}
\end{table*}

\textbf{Dependency analysis}
Dependency grammar is an important tool for natural language understanding. It is able to point out the syntactic collocation relations between words. In this paper, we calculate the dependency properties between individual words and attempt to analyze the differences between human-written and ChatGPT-written abstracts. We compute statistical histograms of dependencies in each type of abstracts, and then rank them according to the percentage. As shown in Figure \ref{fig:dependency}, we present the statistics for the top ten dependencies. There is some similarity in the distribution of the human-written abstracts and ChatGPT-written abstracts, but there are also some differences. Specifically, we have the following findings:

(1) The Adjectival modifier ($amod$), Prepositional modifier ($prep$), and Compound modifier ($compound$) are more used in human-written abstracts. Compared with ChatGPT-written abstracts, human-written abstracts tend to apply more modifiers to define the words precisely.

(2) The proportions of the Direct Object ($dobj$), Nominal subject ($nsubj$), and Root words($ROOT$) in the human-written abstracts are smaller than those of ChatGPT-written abstracts. This is similar with the distributions in the lexical analysis. ChatGPT could provide more the rich vocabulary for synthesizing abstracts.

\begin{figure*}[!h]
	\centering
	\includegraphics[width=1.0\linewidth]{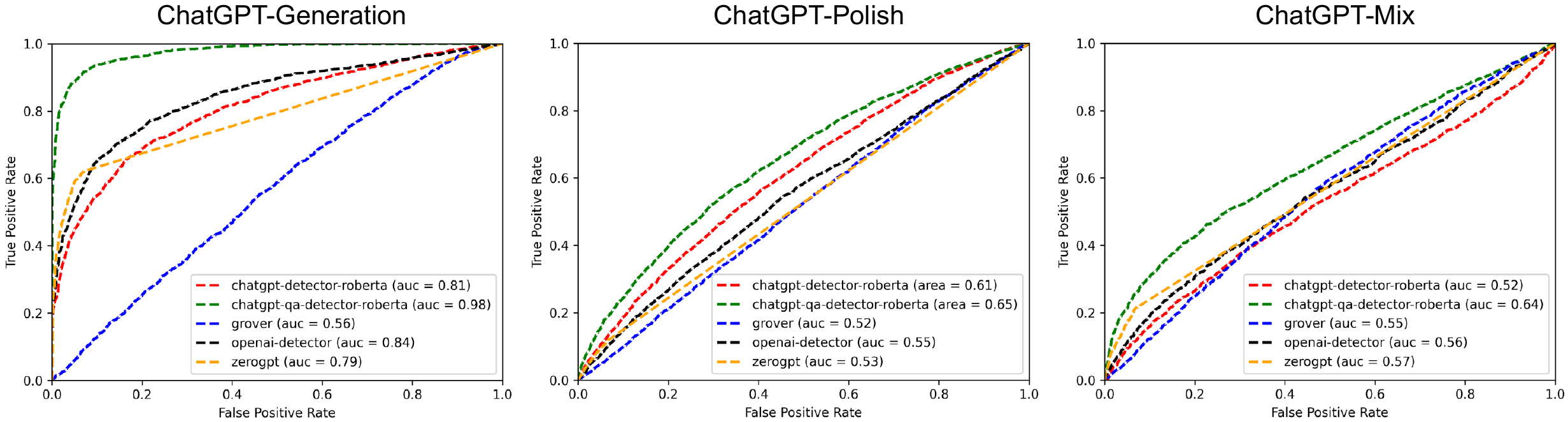}

	\caption{ROC curves of existing detection schemes on three datasets(ChatGPT-Generation, ChatGPT-Polish, and ChatGPT-Mix).}
	\label{fig:roc_withouttrain}
\end{figure*}
\begin{table*}[!h]
	\centering
	\caption{The detection performance of different models. The models are trained on the CHEAT training dataset and then evaluated on the corresponding test dataset.}
	
	\setlength{\tabcolsep}{8.3mm}{
		\begin{tabular}{|c|ccc|}
			\hline
			\multirow{2}{*}{Models}                                                                  & \multicolumn{3}{c|}{Datasets}                                                         \\ \cline{2-4}
			                                                                                         & \multicolumn{1}{c|}{Polish}         & \multicolumn{1}{c|}{Generation} & Mix          \\ \hline
			\begin{tabular}[c]{@{}c@{}}Distilbert \cite{sanh2019distilbert}\end{tabular}           & \multicolumn{1}{c|}{99.43}          & \multicolumn{1}{c|}{100}       & 85.07          \\ \hline
			\begin{tabular}[c]{@{}c@{}}BERT \cite{kenton2019bert}\end{tabular}                     & \multicolumn{1}{c|}{99.48}          & \multicolumn{1}{c|}{100}       & 86.62          \\ \hline
			\begin{tabular}[c]{@{}c@{}}Roberta \cite{liu2019roberta}\end{tabular}                  & \multicolumn{1}{c|}{99.72}          & \multicolumn{1}{c|}{100}       & 52.93          \\ \hline
			\begin{tabular}[c]{@{}c@{}}BERT-multilingual \cite{pires2019multilingual}\end{tabular} & \multicolumn{1}{c|}{99.49}          & \multicolumn{1}{c|}{100}       & 60.16          \\ \hline
			\begin{tabular}[c]{@{}c@{}}PubMedBERT \cite{gu2021domain}\end{tabular}                 & \multicolumn{1}{c|}{\textbf{99.56}} & \multicolumn{1}{c|}{100}       & \textbf{87.83} \\ \hline
		\end{tabular}}
	\label{tab:modeltrain}
\end{table*}
\subsection{Text Synthesis Detection Evaluation}
The distribution differences between ChatGPT-written abstracts and human-written abstracts provide the feasibility of their detection. This subsection evaluate the detection performance of existing algorithms on ChatGPT-written abstracts.

\subsubsection{Compared Text Synthesis Detection Methods}

We consider five text synthesis detection methods in our experiment. Due to the need to evaluate each method on the CHEAT dataset, we only capture those schemes that have code or the model parameters publicly available.
\begin{itemize}
	% \item \textbf{DetectGPT} Mitchell \emph{et al.}\cite{mitchell2023detectgpt} discovered that LLM text tends to occupy the local optimum of the model's log probability function, and the log probability computed when encountering small perturbations is often lower than that of human-written text. Therefore, they defined a curvature-based method for generating text judgment (DetectGPT), which uses the Hessian trace of log probability to distinguish the original text from the generated text. The scheme is able to obtain a highly accurate generative text detector without model training and performs well in news article detection tasks.
	\item \textbf{Grover Detector} Grover is a controlled text generation model proposed by Rowan Zellers et al.\cite{zellers2019defending}. The Grover detector is trained based on the Grover generation model to determine whether the text is generated by the neural network model.
	\item \textbf{ZeroGPT} ZeroGPT detector\cite{AITextDetector} is an online detector, which is trained based on 10 million articles and texts. It is capable of performing machine-generated text detection in multiple languages with high accuracy.
	\item \textbf{OpenAI-detector} The detector\cite{solaiman2019release} is officially provided by OpenAI. It employs the original text and the GPT-2 generated text to fine-tune the RoBERTa model to determine whether the text is machine generated or not.
	\item \textbf{ChatGPT-detector-roberta} Guo \emph{et al.}\cite{guo2023close} constructed the HC3 (Human ChatGPT Comparison Corpus) dataset consisting of nearly 40K questions and their corresponding human/ChatGPT answers, and then used these data to fine-tune the RoBERTa model to obtain an accurate synthetic text detector.
	\item \textbf{ChatGPT-qa-detector-roberta} The correlation between the question and answer is also used to detect the synthetic text.  Guo \emph{et al.}\cite{guo2023close} trained a detection model using Q\&A statements in HC3 dataset, and achieved high accuracy in synthetic text detection.

\end{itemize}
% A brief summary of the models, training datasets, and source code for the various types of generative text detection methods involved is given in Table 3.

\subsubsection{Detection Performance}

In this part, we evaluate the performance of existing detection algorithms on our CHEAT dataset. The detection accuracy (ACC) and area under the ROC curve (AUC) are applied to evaluate the effectiveness of the detection algorithms. We first evaluate the detection performance when the algorithm is not trained with our CHEAT dataset. As shown in Table \ref{tab:compare_withouttrain}, due to the differences of training corpus, existing algorithms have difficulty in detecting ChatGPT-written abstracts. The ROC curve shown in Figure \ref{fig:roc_withouttrain} also demonstrates the weakness of existing detection schemes. The Chatgpt-qa-detector-roberta could obtain high detection accuracy on the ChatGPT-Generation dataset, but fails on the ChatGPT-Polish and ChatGPT-Mix dataset. Existing schemes are still inefficient in detecting ChatGPT-written abstracts.

After that, we trained multiple models using ChatGPT-Generation, ChatGPT-Polish, and ChatGPT-Mix, respectively. The detection performance of trained models is then evaluated on the corresponding test datasets. The AUC scores are presented in Table \ref{tab:modeltrain}. Due to the specialized training on the CHEAT dataset, these models obtain better detection performance on the ChatGPT-written abstracts. In particular, the PubMedBERT, associated with abstract content detection, obtains the best detection performance on our CHEAT dataset. From the results, we believe that the fully generated abstracts are highly feasible to detect. However, similar with the evaluation of existing schemes, the detection difficulty increases with the human involvement. These models could obtain auc scores of 100 on the ChatGPT-Generation dataset while get lower auc score on the ChatGPT-Polish dataset, which are influenced by human writing. Considering the most difficult case, these models obtain the lowest auc score when the human-written abstracts are mixed with ChatGPT-polished abstracts. Detecting ChatGPT-written content with human involvement is still a challenge for existing detection algorithms.
\begin{figure*}[!h]
	\centering
	\includegraphics[width=1.0\linewidth]{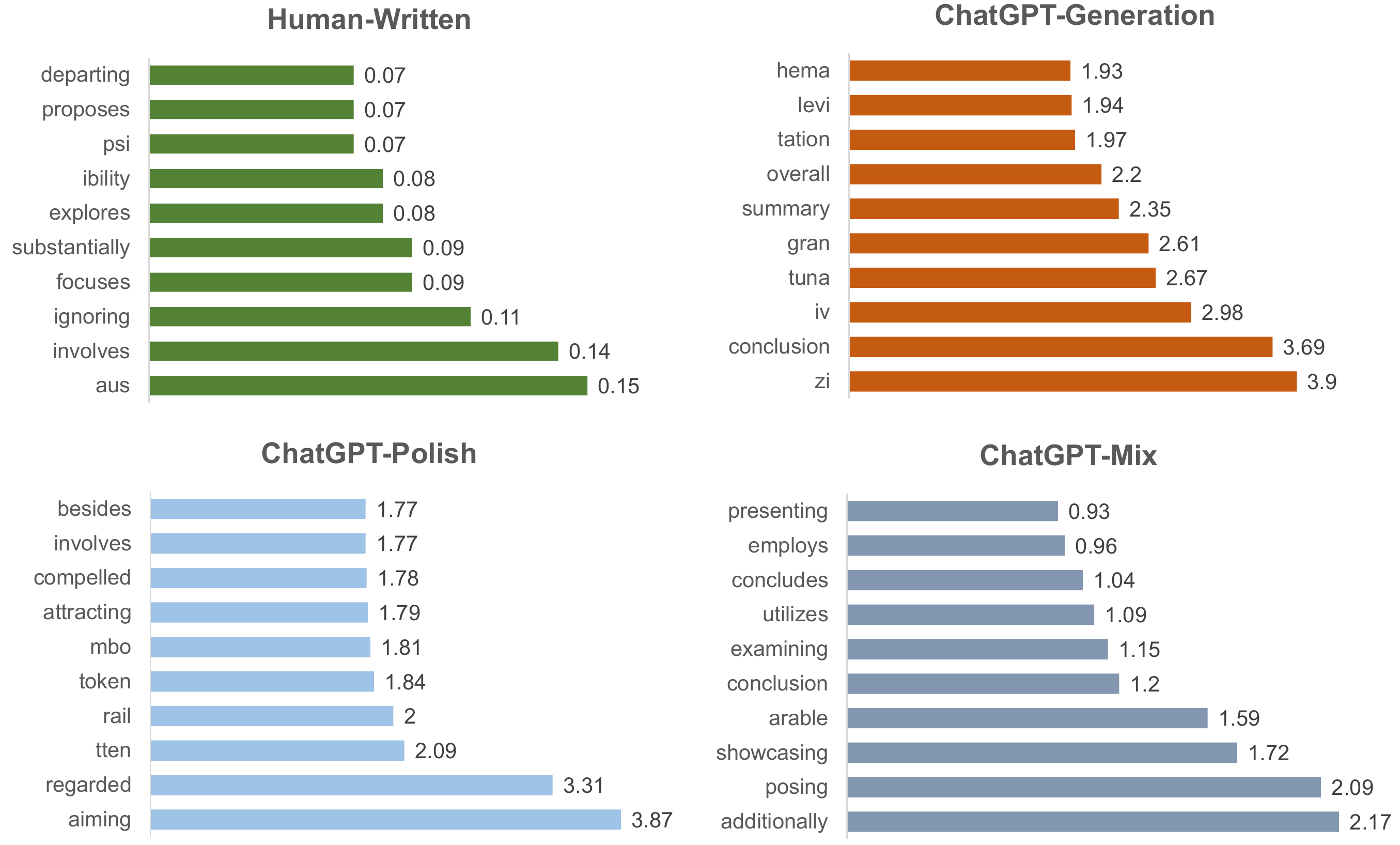}

	\caption{Visualization of SHAP value statistics. The top 10 words ranked by contribution are listed for human-written and ChatGPT-written abstracts.}
	\label{fig:shapvoc}
\end{figure*}
\begin{figure*}[!h]
	\centering
	\includegraphics[width=1.0\linewidth]{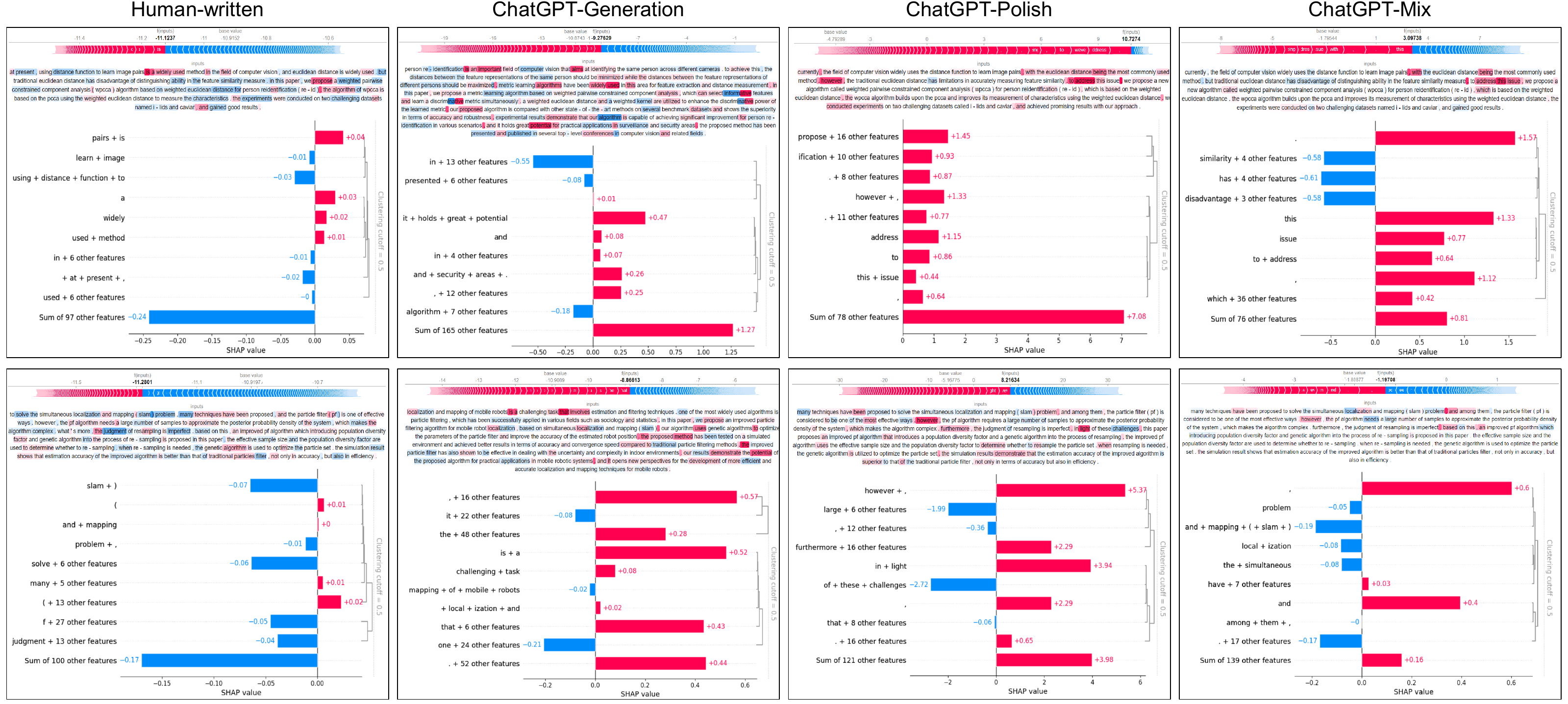}

	\caption{Visualization of SHAP value for single abstract. The sentence semantics and the top 10 words ranked by contribution are calculated for the selected abstract.}
	\label{fig:shapone}
\end{figure*}

\subsubsection{Explainability Analysis}

The well-trained detection model could obtain high accuracy, but have difficulty interpreting the output. It is essential for us to understand whether detection models are perceiving particular word patterns to make judgments. In this subsection, we use SHAP (SHapley Additive exPlanations) \cite{lundberg2017unified} to explain the judgments made by detection models.  Specifically, SHAP uses the classical Shapley values from game theory to link optimal credit allocation with local explanations. It is able to assign feature importance values to each input word of the detection model. To obtain the model interpretion, we first train a high-accuracy detection model based on the PubMedBERT. Then, we compute the SHAP values between the detection results and the input abstracts.  We count the judgment bases for each kind of abstracts, and present the top ten terms of judgment bases in Figure \ref{fig:shapvoc}. When we attempt to discriminate human-written and ChatGPT-generated abstracts, it can be found that ChatGPT tend to use more special-patterns, such as 'zi', 'iv', 'tunv', and 'gran'. When the human involvement increases, precise words gradually play a more important role in detection tasks.  We also visualized the SHAP values for two abstracts in Figure \ref{fig:shapone}. The positive SHAP value indicates a positive impact on classifying the text as ChatGPT-written abstract, and vice versa. It can be seen that detectors could correlate certain words, such as "\emph{it holds great potential}", "\emph{mapping of mobile robots}", and "\emph{and security areas.}", to detect abnormal distributions of ChatGPT-written abstracts. When human involvement increases, words biased towards human-written abstracts would appear, increasing the difficulty of text synthesis detection.

\section{Discussion \& Conclusion}
Although the current ChatGPT demonstrates stunning synthesis ability, we demonstrate that the ChatGPT-written abstracts could be detected by a well-trained detector. To train the detection model using domain-specific knowledge, we construct a large-scale CHatGPT-writtEn AbsTract (CHEAT) dataset, which exceeds the size of all existing publicly available synthetic abstract datasets.

In this paper, we evaluate the performance of existing detection algorithms on the CHEAT dataset. The experimental results in this paper show that \emph{the abstracts generated entirely by ChatGPT are detectable, while the synthetic abstracts with human guidance possess detection difficulty, especially when mixed with human-written text.} We need not be too alarmed by the abstracts fully generated by ChatGPT. Rather, we must direct our attention towards ChatGPT-written abstracts with human guidance. We aspire for this dataset to serve as a stepping stone for ChatGPT-written content detection research, particularly in the realm of paper abstracts.

% \section*{Acknowledgements}

\bibliography{acl2023}
\bibliographystyle{acl_natbib}

\end{document}